\documentclass{bmvc2k}

\title{Multi-Grained Spatio-temporal Modeling for Lip-reading}

\addauthor{Chenhao Wang}{wangchenhao17@mails.ucas.ac.cn}{1}

\addinstitution{
 Institute of Computing Technology, \\
  Chinese Academy of Sciences, Beijing 100190, China
}
\runninghead{Wang}{Multi-Grained Spatio-temporal Modeling for Lip-reading}

\begin{document}

\maketitle

\begin{abstract}
Lip-reading aims to recognize speech content from videos via visual analysis of speakers' lip movements. This is a challenging task due to the existence of \textit{homophemes} -- words which involve identical or highly similar lip movements, as well as diverse lip appearances and motion patterns among the speakers. %
To address these challenges, we propose a novel lip-reading model which captures not only the nuance between words but also styles of different speakers, by a multi-grained spatio-temporal modeling of the speaking process.
Specifically, we first extract both frame-level fine-grained features and short-term medium-grained features by the visual front-end, which are 
then combined to obtain discriminative representations for words with similar phonemes. Next, a bidirectional ConvLSTM augmented with temporal attention aggregates spatio-temporal information in the entire input sequence, which is expected to be able to capture the coarse-gained patterns of each word and robust to various conditions in speaker identity, lighting conditions, and so on. By making full use of the information from different levels in a unified framework, the model is not only able to distinguish words with similar pronunciations, but also becomes robust to appearance changes. We evaluate our method on two challenging word-level lip-reading benchmarks and show the effectiveness of the proposed method, which also demonstrate the above claims. 
\end{abstract}

\section{Introduction}
\label{sec:intro}

Lip-reading, the ability to understand speech using only visual information, is an attractive but highly challenging skill. It plays a crucial role in human communication and speech understanding, as highlighted by the McGurk effect. There are several valuable applications, such as aids for hearing-impaired or speech-impaired persons, analysis of silent movies, and liveness verification in video authentication systems. It is also an important complement to the acoustic speech recognition systems, especially in noisy environments. For such reasons and also the development of deep learning which enables efficient feature learning and extraction, lip-reading has been receiving more and more attention in recent years.

A typical lip-reading framework consists of two steps: analyzing the motion information in the image sequence, and converting that information into words or sentences. One common challenge in this process is various imaging conditions, such as poor lighting, strong shadows, motion blur, low resolution, foreshortening, etc. More importantly, there is a fundamental limitation on performance due to \textit{homophemes}. These are many words or phrases that sound different, but involve the same or very similar movements of the speaker's lips. For example, the phonemes "p", "b" in English are visually identical; while the words "pack" and "back", are homophemes that can hardly be distinguished through lip-reading when there is no more context information.

Motivated by these problems, we hope to build a model which utilizes both fine-grained and coarse-grained spatio-temporal features to enhance the model's discriminative power and robustness. Specifically , we propose a multi-grained spatio-temporal network for lip-reading. The front-end network uses a spatio-temporal ConvNet and a spatial-only ConvNet in parallel, which extract  medium-grained short-term and fine-grained, per-time-step features respectively. In order to fuse these features more effectively, we introduce a spatial attention mask to learn an adaptive, position-wise feature fusion strategy. A two-layer bidirectional ConvLSTM augmented with (forward) input attention is used as the back-end to generate the coarse-grained long-term spatio-temporal features.

In summary, we make three contributions. Firstly, we propose a novel multi-grained spatio-temporal network to solve the lip-reading problem. Secondly, instead of simple concatenation, we fuse the information of different granularity with a learnable spatial attention mechanism. Finally, we apply ConvLSTM to the lip-reading task for the first time. We report the word classification results on two challenging lip-reading datasets, LRW and \textit{LRW}-1000.

\section{Related Work}

In this section, we briefly summarize previous related work about lip-reading and ConvLSTMs.
\vspace{-1.5em}
\paragraph{Lip reading.}Research on lip-reading has a long history. Most early methods are based on carefully hand-engineered features. A classical type of methods is to use Hidden Markov Models (HMM) to model the temporal structure within the extracted frame-wise features \cite{chiou1997lipreading,potamianos2003recent,chandrasekaran2009natural}. Other well-known features include the Discrete Cosine Transform (DCT) \cite{potamianos1998image}, Active Appearance Model (AAM), Motion History Image (MHI) \cite{duchnowski1995toward}, Local Binary Pattern (LBP) \cite{zhao2009lipreading}, and vertical optical flow based features \cite{shaikh2010lip}. With the rapid development of deep learning technologies and the appearance of large-scale lip-reading databases \cite{chung2016lip,1chung2017lip,yang2018lrw}, researchers have started to use convolutional neural networks to extract the features of each frame and also use recurrent units for holistic temporal modeling \cite{noda2015audio,thangthai2015improving,almajai2016improved}. In 2016, \cite{chung2016lip} proposed the first large-scale word-level lip-reading database together with several end-to-end lip-reading models. Since then, more and more work perform end-to-end recognition with the help of deep neural networks (DNN).

According to the design of the front-end network, these modern lip-reading methods can be roughly divided into three categories: (a) fully 2D CNN based, which build on the success of 2D ConvNets in image representation learning;  
(b) fully 3D CNN based, which is inspired by the success of 3D ConvNets in action recognition, among which LipNet\cite{assael2016lipnet} is a representative work that yields good results on the GRID audiovisual corpus; and (c) mixture of 2D and 3D convolutions, which inherit the merits of both (a) and (b) by capturing the temporal dynamics in a sequence and extracting discriminative features in the spatial domain simultaneously. Recently, methods of type (c) have become dominant in lip-reading due to its excellent performance. For example, in 2018, \cite{1petridis2018end} attained $83.0$\% word accuracy on the LRW dataset based on the type (c) architecture, achieving a new state-of-the-art result. However, the above method simply stacks 3D and 2D convolutional layers, which may not fully unleash the power of the two components. Our model proposes a new approach to take the respective advantages of 3D and 2D ConvNets, by using them as two separate branches and fusing the features adaptively, similar to the popular two-stream architecture for action recognition \cite{simonyan2014two}.
\vspace{-1.5em}
\paragraph{LSTM and ConvLSTM.}For general-purpose sequence modeling, LSTM \cite{hochreiter1997long} as a special RNN structure has been proven stable and powerful in modeling long-range dependencies. LSTMs often lead to better performance where temporal modeling capacity is required, and are thus widely used in NLP, video prediction, lip-reading, and so on. A common practice of using LSTMs in video recognition is to employ a fully-connected layer before the LSTM. Although this FC-LSTM layer has been proven powerful for temporal modeling, it loses too much information about the spatial correlation in the data. To address this, Shi et al. proposed ConvLSTM\cite{Shi2015Convolutional}, which is based on the LSTM design but considers both temporal and spatial correlation in a video sequence with additional convolution operations, effectively fusing temporal and spatial features. It has been successfully applied to action recognition \cite{li2018videolstm,wang2018human}, gesture recognition \cite{zhu2017multimodal,1zhang2017learning} and other fields \cite{sudhakaran2017convolutional}. Additionally, a new spatio-temporal LSTM unit \cite{1wang2017predrnn} is recently designed to memorize both temporal and spatial representations, obtaining better performance than the conventional LSTM.

In this paper, we introduce ConvLSTM to the lip-reading task for the first time. When aggregating information from the whole lip sequence, its ability to capture both long and short term temporal dependencies while considering the spatial relationships in feature maps makes it ideal for accommodating to differences across speakers. We also augment the ConvLSTM with an attention mechanism on the inputs, which will be described in detail in Sec.~\ref{sec:attconvlstm}

\vspace{-1em}
\section{Multi-Grained Spatio-temporal Modeling For Lip-reading}
\begin{figure*}
  \centering
  \includegraphics[width=1\linewidth]{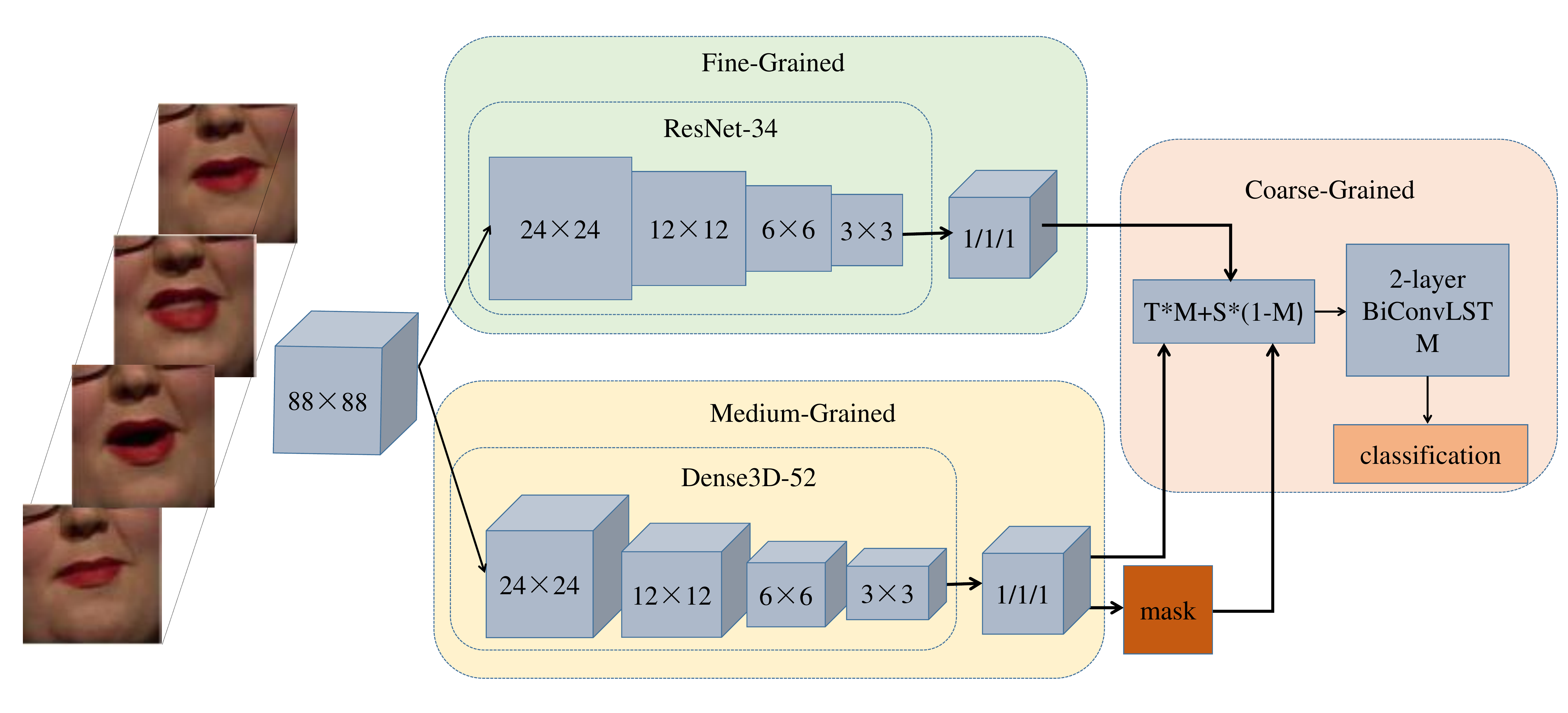}
\caption{The architecture of the proposed framework, which consists of a spatio-temporal convolution module followed by a two-branch structure and a two-layer Bi-ConvLSTM with forward input attention. Finally, a fully-connected layer is used to obtain the prediction results.}
\label{fig:whole}
\vspace{-1em}
\end{figure*}
Given a sequence of the mouth region corresponding to an utterance, our goal is to capture both the fine-grained patterns that can distinguish one word from another, and the coarse-grained patterns describing mouth shapes and motion information that are ideally invariant to the varied styles of different speakers.

As mentioned earlier, simply cascading 2D and 3D convolution may not be optimal for lip-reading, since some movements may be very weak and thereby lost during pooling. Therefore, we split the learning process into three sub-networks that complement each other. In this section, we present the proposed multi-grained spatio-temporal framework which learns the latent spatio-temporal patterns of different words from three different spatio-temporal scales for the lip-reading task. As shown in Fig.~\ref{fig:whole}, the network consists of a 2D ResNet-34 based fine-grained module, a 52-layer DenseNet-3D medium-grained module, and a coarse-grained module that adaptively fuses and aggregates the features from these two modules. By jointly learning the latent patterns at multiple spatio-temporal scales and efficiently fusing these information, we achieve much better performance. We now give a detailed description of the architecture.
\vspace{-1em}
\subsection{Fine-grained Module}
Words with similar mouth movements are fairly ubiquitous. However, when we compare the sequences side by side and examine each time-step, very often we can still observe slight differences in appearance. This observation leads to the idea that enhancing spatial representations alone to some extent may improve the discriminative power of the model. As an effective tool to capture the salient features in images, 2D convolutional operations have been proven successful in several related tasks, such as image recognition, object detection, segmentation, and so on. We introduce cascaded 2D convolutional operations here to extract the salient features in each frame. Different from the traditional role of 2D convolutional operation in other methods, the 2D convolutions introduced here should not merely function as a feature extractor, but highlight salient appearance cues in each frame, which will eventually help enhance the fine-grained patterns for subtle differences among words. In our model, the 2D ConvNet is a 34-layer ResNet.

\vspace{-1em}
\subsection{Medium-grained Module}
3D convolution have become widely adopted in video recognition and proven capable of capturing short-term spatio-temporal patterns. They are expected to be more robust than using 2D convolutions which produce frame-wise features because they account for motion information. Moreover, while there are words with subtle differences that require fine-grained information, most words are still able to be distinguished through the ways they are pronounced, albeit somewhat speaker-dependent. This requires the model to be capable of modeling medium-grained, short-term dynamics, which is a job suitable for 3D convolutions. In our model, the medium-grained 3D ConvNet is a 52-layer 3D-DenseNet \cite{yang2018lrw}.

\vspace{-1em}
\subsection{Coarse-grained Module}
The coarse-grained module begins by fusing the features from the previous two modules. Different from most previous methods which directly cascade 2D and 3D convolutions, we introduce an attention mechanism to combine the fine-grained features and the medium-grained features into a primary representation. As shown in Fig.~\ref{fig:whole}, the attention mask is implemented with an $1\times1\times1$ convolution, which adaptively adjusts the fusion weights at each spatial location. This spatial attention mask and the final fused features $\mathbf{F}$ are obtained by
\begin{equation}
    \begin{aligned}
    \mathbf{S} = \mathtt{2DCNN}(\mathbf{X}), &\quad \mathbf{T} = \mathtt{3DCNN}(\mathbf{X}), \\
    \mathtt{mask} &=\sigma(\mathbf{W}\mathbf{T}),\\
    \mathbf{F} = \mathbf{T} \odot \mathtt{mask} &+ \mathbf{S}\odot (1-\mathtt{mask}).
    \end{aligned}
\end{equation}
where $\mathbf{X}$ are the input feature maps, $\mathbf{S}$, $\mathbf{T}$ are the respective outputs of the two branches, $\mathbf{W}$ is a learned parameter, $\sigma$ is the sigmoid function, and $\odot$ denotes element-wise multiplication.

\begin{figure*}
  \centering
  \centerline{\includegraphics[width=0.5\linewidth]{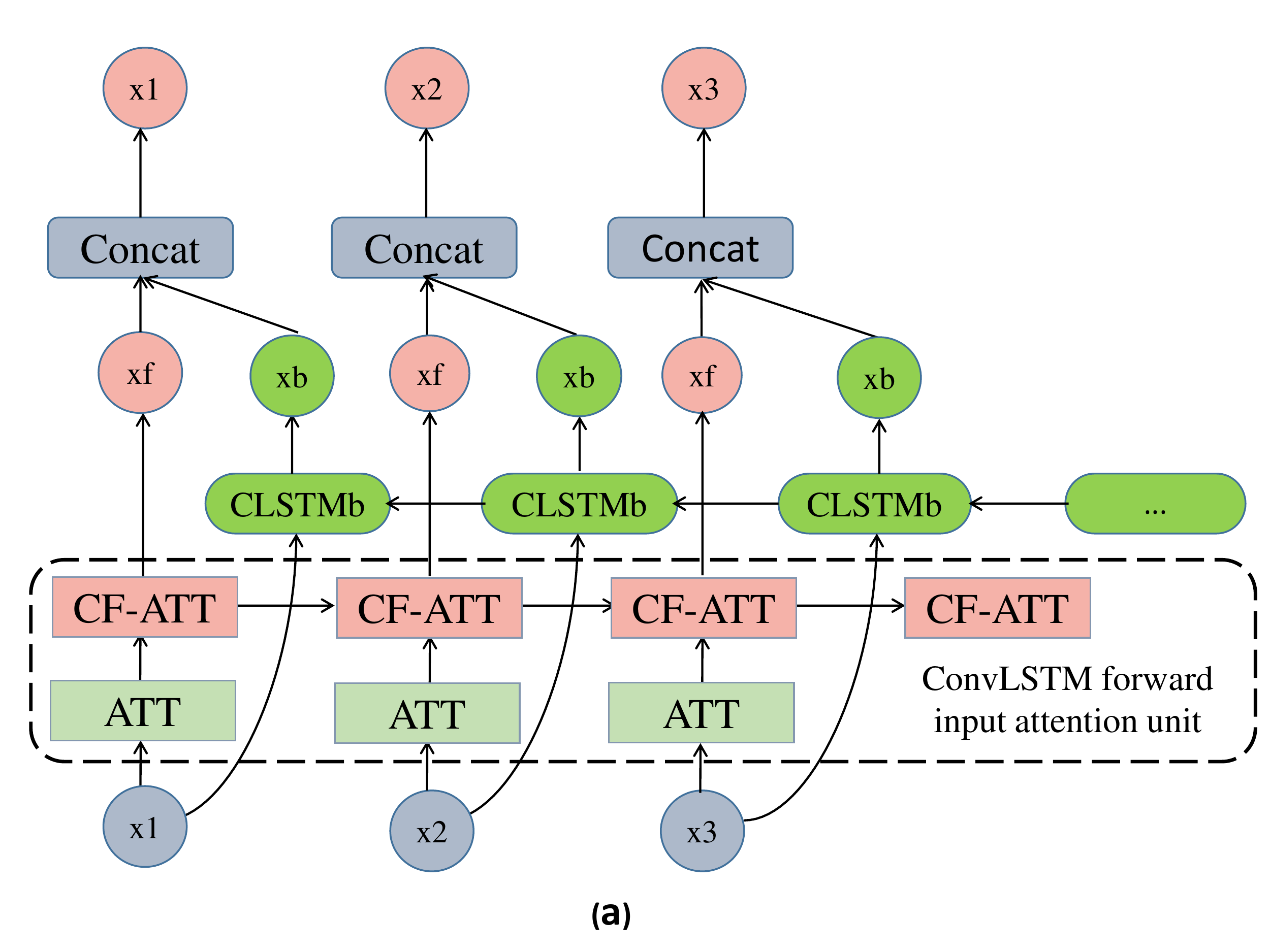}\includegraphics[width=0.5\linewidth]{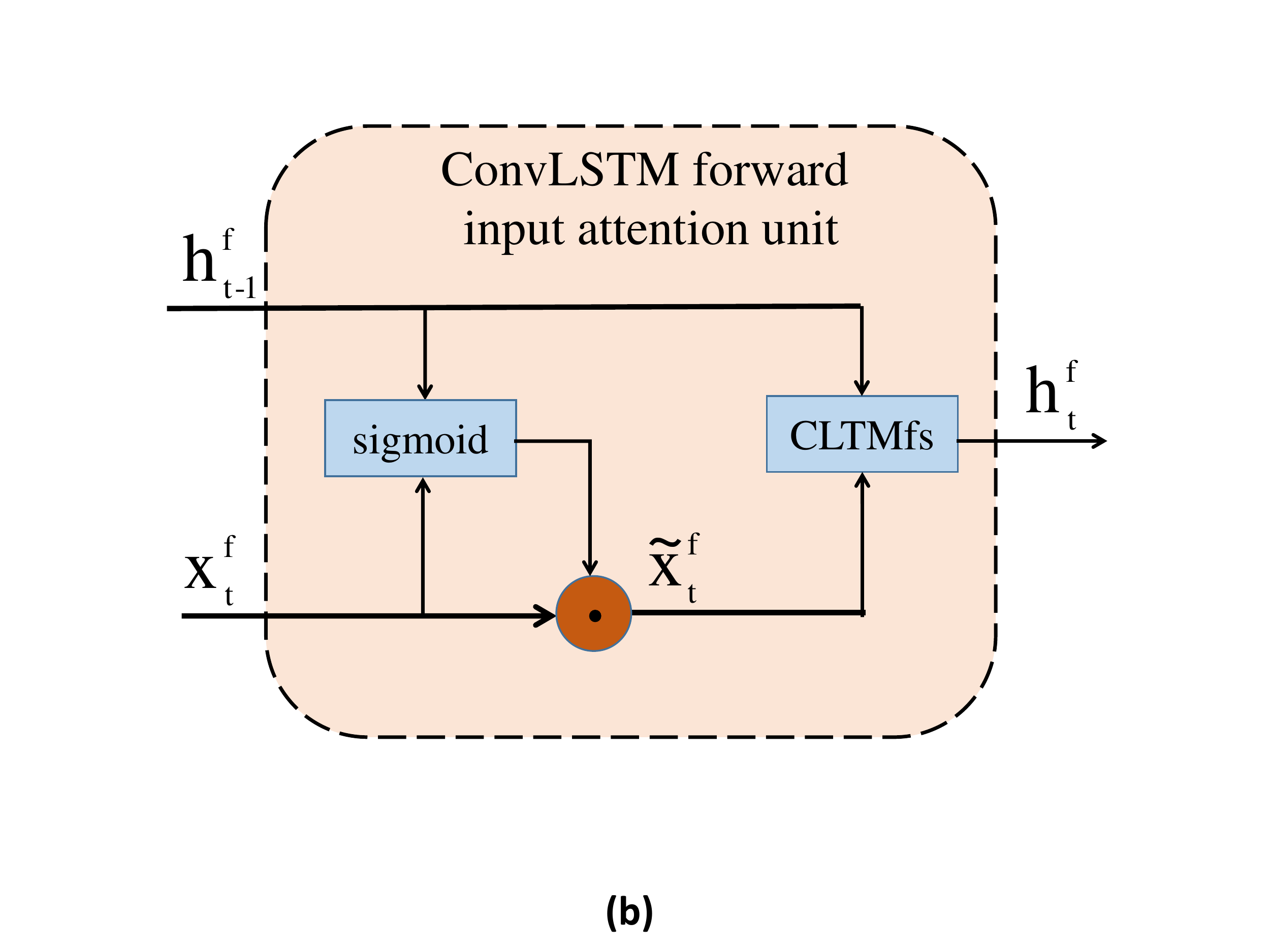}}
\caption{(a) The forward input attention augmented Bi-ConvLSTM. The attention-augmented ConvLSTM layer CF-ATT processes the inputs in the forward direction attentively, while the plain ConvLSTM layer CLSTMb processes information in reversed order. (b) The ConvLSTM forward input attention unit, where $\odot$, as before, denotes element-wise multiplication.}
\label{fig:convlstm}
\end{figure*}
Every person has his or her own speaking style and habits, such as nodding or turning his or her head while speaking. Meanwhile, owing to the appearance factors such as lighting conditions, speaker's pose, make-up, accent, age and so on, the image sequences of even the same word would have several different styles. Considering the diversity of the appearance factors, a robust lip-reading model has to model the global latent patterns in the sequence in a high-level to highlight the representative patterns and cover the slight style-variations in the sequence.
FC-LSTMs are capable of modeling long-range temporal dependencies and have a powerful gating mechanism. But the major drawback of FC-LSTM in handling spatio-temporal data is its usage of full connections in input-to-state and state-to-state transitions in which no spatial correlation is encoded. To overcome this problem, we use a two-layer bidirectional ConvLSTM module augmented with forward input attention which proceeds to model the global latent patterns in the whole sequence based on the fused initial representations. It is able to cover the various styles and speech modes in the speaking process, which will be demonstrated in the experiment section.

\subsubsection{Bi-ConvLSTM with Forward Input Attention}
\label{sec:attconvlstm}
Compared with the conventional LSTM, ConvLSTM proposed in \cite{Shi2015Convolutional}, as a convolutional counterpart of conventional fully connected LSTM, introduces the convolution operation into input-to-state and state-to-state transitions. ConvLSTM is capable of modeling 2D spatio-temporal image sequences by explicitly encoding their 2D spatial structures into the temporal domain. ConvLSTM models temporal dependency while preserving spatial information. Thus it has been widely applied for many spatio-temporal  tasks. Similar to FC-LSTM, a ConvLSTM unit consists of a memory cell  $\mathbf{c _ { t }}$, an input gate  $\mathbf{i _{ t }}$, an output gate  $\mathbf{o _{ t }}$ and a forget gate  $\mathbf{f_{ t }}$. The main equations of ConvLSTM are as follows:
\begin{equation}
    \begin{aligned}
    \mathbf{i}_{t} &=\sigma\left(\mathbf{W}_{x i} * \mathbf{X}_{t}+\mathbf{W}_{h i} * \mathbf{H}_{t-1}+\mathbf{W}_{c i} \circ \mathbf{C}_{t-1}+\mathbf{b}_{i}\right) \\
    \mathbf{f}_{t} &=\sigma\left(\mathbf{W}_{x f} * \mathbf{X}_{t}+\mathbf{W}_{h f} * \mathbf{H}_{t-1}+\mathbf{W}_{c f} \circ \mathbf{C}_{t-1}+\mathbf{b}_{f}\right) \\
    \mathbf{C}_{t} &=\mathbf{f}_{t} \circ \mathbf{C}_{t-1}+\mathbf{i}_{t} \circ \tanh \left(W_{x c} * \mathbf{X}_{t}+\mathbf{W}_{h c} * \mathbf{H}_{t-1}+\mathbf{b}_{c}\right) \\
    \mathbf{o}_{t} &=\sigma\left(\mathbf{W}_{x o} * \mathbf{X}_{t}+\mathbf{W}_{h o} * \mathbf{H}_{t-1}+\mathbf{W}_{c o} \circ \mathbf{C}_{t}+\mathbf{b}_{o}\right) \\
    \mathbf{H}_{t} &=\mathbf{o}_{t} \circ \tanh \left(\mathbf{C}_{t}\right)
    \end{aligned}
\end{equation}
where `$\ast$' denotes the convolution operator and `$\circ$' denotes the Hadamard product. 

However, the structures of existing RNN neurons mainly focus on controlling the contributions of current and historical information but do not explore the difference in importance among different time-steps \cite{1zhang2018adding}. So we introduce an attention mechanism to the forward direction of the bidirectional ConvLSTM, as shown in Fig.~\ref{fig:convlstm}. The input attention can determine the relative importance of different frames and assign a suitable weight to each time-step. This augmented Bi-ConvLSTM can not only learn spatial temporal features but also select important frames. We only use attention on the inputs to Bi-ConvLSTM's forward direction:
\begin{equation}
    \mathbf{a}_t=\sigma(\mathbf{W}_{\mathbf{X} \mathbf{a}}\mathbf{X}_{\mathbf{f}; t}+\mathbf{W}_{\mathbf{h a}}\mathbf{h}_{\mathbf{f}; t-1})
\end{equation}
where the current (forward) input $\mathbf{X}_{\mathbf{f}; t}$ and the previous hidden state $\mathbf{h}_{\mathbf{f}; t-1}$ are used to determine the levels of importance of each frame of the forward input  $\mathbf{X}_{\mathbf{f}; t}$ .

The attention response modulates the forward input and computes
\begin{equation}
    \widetilde{\mathbf{X}}_{\mathbf{f}; t}=\mathbf{a _ { t }}\circ \mathbf{X}_{\mathbf{f}; t}
\end{equation}
The recursive computations of activations of the other units in the RNN block are then based on the attention-weighted input $\widetilde{\mathbf{X}}_{\mathbf{f}; t}$, instead of the original input $\mathbf{X}_{\mathbf{f}; t}$.

\section{Experiments}
In this section, we present the results of our experiments on the word-level LRW and LRW-1000 datasets. We give a brief description to the two datasets and our implementation, and finally a detailed analysis of our experimental results.
\vspace{-1em}
\subsection{Datasets}
\paragraph{Lip Reading in the Wild (LRW) \cite{chung2016lip}.}The LRW database consists of short segments ($1.16$ seconds) from BBC programs, mainly news and talk shows. It is a very challenging dataset since it contains more than $1000$ speakers and large variations in head pose and illumination. For each target word, it has a training set of $1000$ segments, a validation and an evaluation set of $50$ segments each. The total duration of this corpus is $173$ hours. The corpus with $500$ words is also much larger than previous lip-reading databases used for word recognition.

\paragraph{\textit{LRW}-1000 \cite{yang2018lrw}.}\textit{LRW}-1000 is a challenging Mandarin lip-reading dataset due to its large variations in scale, resolution, background clutter, and speaker attributes. The speakers are mostly interviewers, broadcasters, program guests, and so on. The dataset consists of $1000$ word classes and has $718,018$ samples, totaling $57$ hours. The minimum and maximum length of the samples are about 0.01 seconds and 2.25 seconds respectively, with an average of about 0.3 seconds for each sample.

\begin{figure*}
  \centering
  \centerline{\includegraphics[width=1\linewidth]{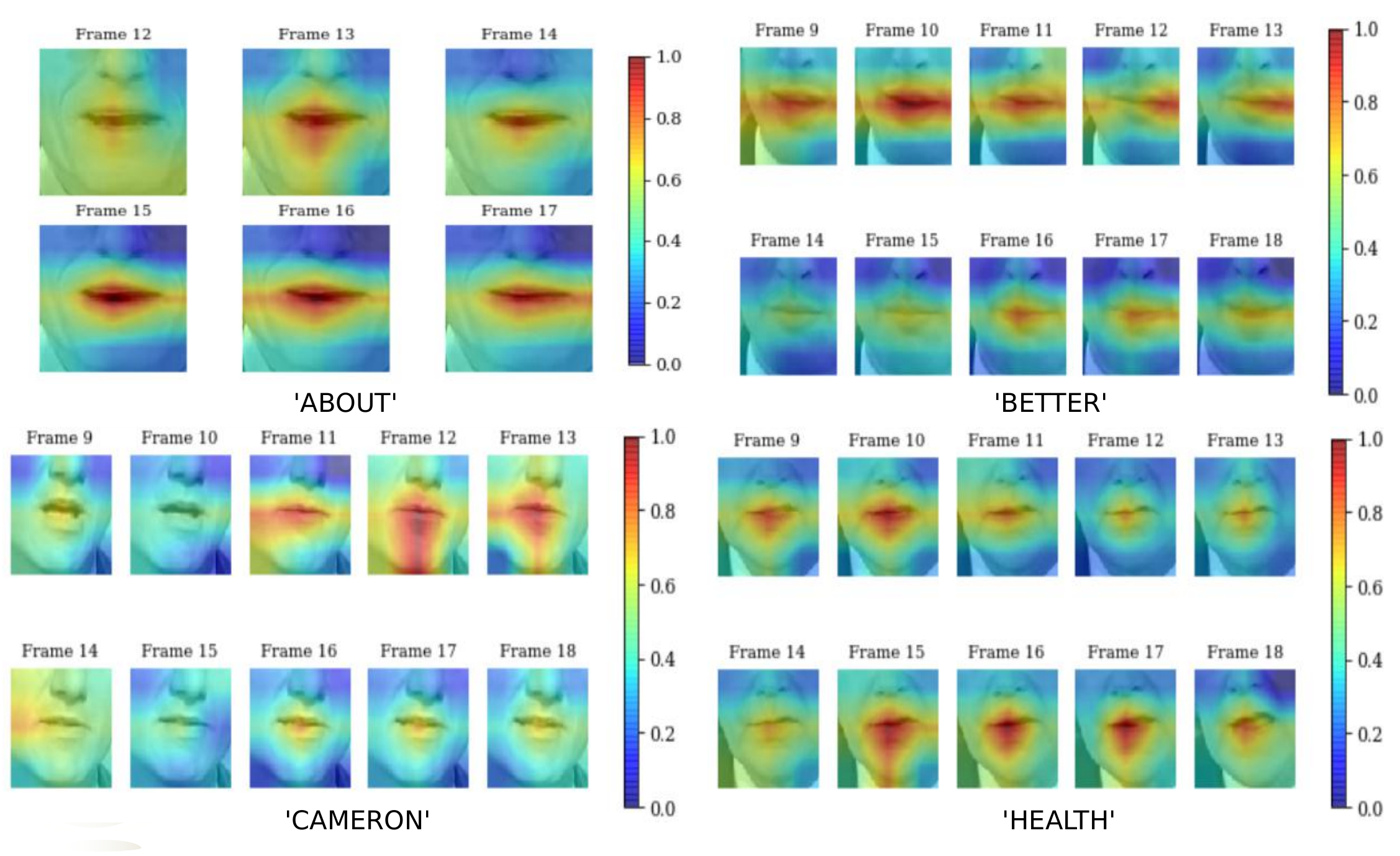}}
\caption{The attention mask automatically adjusts the position-specific fusion weights and generates an initial representation. For clarity, only the frames corresponding to the target word are shown.}
\label{fig:mask}
\vspace{-1em}
\end{figure*}
\vspace{-1em}
\subsection{Implementation Details}
Our models are implemented with PyTorch and trained on servers with three NVIDIA Titan X GPUs, each with 12GB memory. In our experiments, for the LRW dataset, the mouth region of interests (ROIs) are already centered, and a fixed bounding box of 96 by 96 is used for all videos. All images are converted to grayscale, and then cropped to $88\times 88$. As an effective data augmentation step, we also randomly flip all the frames in the same sequence horizontally. For the two-branch models, we first train each individual branch to convergence, and then fine-tune the model end-to-end. We use the Adam optimizer with an initial learning rate of $0.0001$ and a momentum of $0.9$. During the fine-tuning with RGB LRW-1000, the maximum number of frames is set to $30$.

The first convolutional layer has kernel of size $64\times5\times7\times7$ (channels / time / height / width), while max pooling has a kernel of size $1\times3\times3$. We then reshape the feature map to $24\times24$.
In our model, the two branches are constructed by
 a $34$-layer ResNet and a $52$-layer 3D-DenseNet \cite{yang2018lrw} respectively. We use a $1\times1\times1$ 3D convolution to reduce the dimensionality. Then $512\times29\times3\times3$ fusion feature is fed to a two-layer Bi-ConvLSTM with forward input attention. The Bi-ConvLSTM has kernel size $3\times3$. The output layer is a fully connection layer to obtain prediction results. We average the framewise prediction for the final results. The two blocks of layers transform the tensors as $88\times88 \rightarrow 22\times22 ~\overrightarrow{\text{\footnotesize upsample}}~ 24\times24 \rightarrow 12\times12 \rightarrow 6\times6 \rightarrow 3\times3$.
\subsection{Results}
Performance estimates are expressed in terms of word-level error rate on LRW dataset and \textit{LRW}-1000 dataset, respectively. We set up a few control experiments including only 2D CNN branch, only 3DCNN branch, two-branch / Bi-GRU, two-branch / Bi-ConvLSTM and our model. Results on two datasets are provided in Table~\ref{tab:lrw-results}. On the LRW dataset, our model shows marginally better results
which we believe is because the model can learn the multi-grained spatio-temporal features. 

\begin{table}[]
\caption{Classification accuracy of the two-branch network on the LRW database and LRW-1000 database. `ResNet-34' uses the 34-layer ResNet frontend proposed in \cite{1petridis2018end} (the results are our reproduction). `DenseNet-3D' uses a 52-layer DenseNet-3D front-end proposed in \cite{yang2018lrw}.}
\centering
 \vspace{+1em}
\begin{tabular}{|c|c|c|}
\hline
\textbf{Method}                & \textbf{LRW}        & \textbf{\textit{LRW}-1000} \\ \hline
DenseNet-3D + Bi-GRU     & $81.70$\%          & $34.76$\% \cite{yang2018lrw}       \\ \hline
ResNet-34 + Bi-GRU       & $81.70$\%          & $\mathbf{38.19}\%$ \cite{yang2018lrw}        \\ \hline
Two-branch + Bi-GRU      & $82.98$\%          & $36.48$\%            \\ \hline
Two-branch + Bi-ConvLSTM & $83.15$\%          & $36.12$\%            \\ \hline
\textbf{Proposed Model}    & $\mathbf{83.34\%}$ & $36.91\%$   \\ \hline
\end{tabular}
\label{tab:lrw-results}
\vspace{-1em}
\end{table}

From Table~\ref{tab:lrw-results} we can find that the ResNet-34 model and the DenseNet-3D model perform equally well on the LRW dataset, achieving an accuracy of $81.70$\%. However, the recognition results of these two structures are different. In LRW-1000, the ResNet-34 / Bi-GRU is better than 3D-DenseNet / Bi-GRU. The possible reason for this is that 2D CNN can better capture the fine-grained features in each time-step to discriminate words. Compared with the baseline two-branch models, we introduce the soft attention based fusion mechanism to learn an adaptive weight to keep the most discriminative information from the two branches and indeed to lead to more powerful spatio-temporal features. On LRW dataset, compared with the results of our ResNet-34 + Bi-GRU baseline, there is an increase of 1.28\%. But the two-branch performance is higher than the DenseNet-3D / Bi-GRU results. The attention mask is shown in Fig.~\ref{fig:mask}. From these figures, we can find that the attention mask can learn the weights well. It can pay close attention to the lip area to make the learning process automatically modify the fusion weights to generate the early-stage representation. Therefore the two-branch / Bi-GRU architecture can obtain more robust results.

For the LRW database, compared with two-branch / Bi-GRU and two-branch / Bi-ConvLSTM, it is clear from the results that bidirectional ConvLSTM modules are able to significantly improve the performance over two-branch / Bi-GRU. This structure not only indicates that temporal information has been learned but also highlights the importance of spatial information for the lip-reading task.  

Clips from the LRW dataset include context and may introduce redundant information to the network. From Table~\ref{tab:sota} we can find that the Bi-ConvLSTM with forward input attention works better, likely because it can focus on controlling the contributions of current and historical different importance levels on different frames and identify the most important ones. Table~\ref{tab:sota} shows the effectiveness of our forward input attention Bi-ConvLSTM. Therefore our model outperforms the two-branch / Bi-ConvLSTM.

\subsection{Comparison with the state-of-the-art}
Table~\ref{tab:sota} summarizes the performance of state-of-the-art networks on LRW and LRW-1000. Our network has an absolute increase of 1.6\% over our reproduction of the baseline ResNet-34 model in \cite{1petridis2018end} on LRW database. From the above results we see that the mixed 3D-2D architecture still shows very strong performance. However the results also shows the importance of fine-grained spatio-temporal features in the lip-reading task. 
The results also confirm that it is reasonable to use the attention mask to merge the fine-grained and medium-grained features, and replace FC-LSTM with ConvLSTM. Our model takes full advantage of the 3D ConvNet, the 2D ConvNet and the ConvLSTM. The proposed attention-augmented variant of ConvLSTM further enhances its ability for spatio-temporal feature fusion. The forward input attention in Bi-ConvLSTM not only learns spatial and temporal features but also explore the different importance levels of different frames. But we reproduced the 3D+2D model in the database of the accuracy is lower in the \cite{yang2018lrw}. This reason may be that we do not use the fully-connected layers in the model, and we also do not use three-stage training. Therefore, the best recognition results can be obtained by taking full use of the intrinsic advantages of the different networks. 

\begin{table}
\caption{Comparison with the state-of-the-art on the test set of LRW and LRW-1000. '(reproduced)' denotes the result of our reproduction.}  
\centering
\subtable[the state-of-the-art on the LRW]{  
       \begin{tabular}{|c|c|}
\hline
\textbf{Method}             & \textbf{Accuracy}                   \\ \hline
Chung18 \cite{chung2018learning}            & $71.50$\%                    \\ \hline
Chung17 \cite{1chung2017lip}              & $76.20$\%                     \\ \hline
Petridis18 (end-to-end) \cite{1petridis2018end}       & $82.00$\%                     \\ \hline
Petridis18 (reproduced)          & $81.70$\% \\ \hline
Stafylakis17 \cite{stafylakis2017combining}        & $83.00$\%                     \\ \hline

\textbf{Proposed Model} & $\mathbf{83.34\%}$            \\ \hline
\end{tabular} 
       \label{tab:firsttable}  
} 
\qquad  
\subtable[the state-of-the-art on LRW-1000]{          
       \begin{tabular}{|c|c|}
\hline
\textbf{Method}             & \textbf{Accuracy}                   \\ \hline
LSTM-5 \cite{yang2018lrw}            & $25.76$\%                    \\ \hline
D3D \cite{yang2018lrw}              & $34.76$\%                     \\ \hline
3D+2D \cite{yang2018lrw}       & $38.19$\%                     \\ \hline
3D+2D (reproduced)          & $33.78$\% \\ \hline

\textbf{Proposed Model} & $\mathbf{36.91\%}$            \\ \hline
\end{tabular}  
} 
\label{tab:sota}
\vspace{-2em}
\end{table}

\section{Conclusion}

\label{sec:conclusion}

We have proposed a novel two-branch model with forward input attention augmented Bi-ConvLSTM for lip-reading. The model utilizes both 2D and 3D ConvNets to extract both frame-wise spatial features and short-term spatio-temporal features, and then fuses the features with an adaptive mask to obtain strong, multi-grained features. Finally, we use a Bi-ConvLSTM augmented with forward input attention to model long-term spatio-temporal information of the sequence. Using this architecture, we demonstrate state-of-the-art performance on two challenging lip-reading datasets. We believe the model has great potential beyond visual speech recognition. How to better utilize spatial information in temporal sequence modeling to obtain more fine-grained spatio-temporal features is also a worthwhile research. In the future, we will continue to simplify the front-end and extract multi-grained features with a more lightweight structure.

\bibliography{egbib}
\end{document}